\documentclass[letterpaper]{article} 
\usepackage{aaai2026}  
\usepackage{times}  
\usepackage{helvet}  
\usepackage{courier}  
\usepackage[hyphens]{url}  
\usepackage{graphicx} 
\usepackage{natbib}  
\usepackage{caption} 
\usepackage{algorithm}
\usepackage{algorithmic}

\usepackage{multirow}
\usepackage{amsmath}
\usepackage{booktabs}
\usepackage{tabularx}
\usepackage{enumerate}
\usepackage{array}

\newcommand{\myfig}[1]{Figure~\ref{#1}}

\newcommand{\mytable}[1]{Table~\ref{#1}}
\newcolumntype{C}{>{\centering\arraybackslash}X}
\newcolumntype{L}{>{\arraybackslash}X}

\title{PCoKG: Personality-aware Commonsense Reasoning with Debate}
\author {
    Weijie Li\textsuperscript{\rm 1}, Zhongqing Wang\textsuperscript{\rm 1}\thanks{Corresponding Author}, Guodong Zhou\textsuperscript{\rm 1}
}
\affiliations {
    \textsuperscript{\rm 1}School of Computer Science and Technology, Soochow University\\
    silverbeats@qq.com, wangzq@suda.edu.cn, gdzhou@suda.edu.cn
}

\begin{document}

\maketitle

\begin{abstract}
Most commonsense reasoning models overlook the influence of personality traits, limiting their effectiveness in personalized systems such as dialogue generation. To address this limitation, we introduce the Personality-aware Commonsense Knowledge Graph (PCoKG), a structured dataset comprising 521,316 quadruples. We begin by employing three evaluators to score and filter events from the ATOMIC dataset, selecting those that are likely to elicit diverse reasoning patterns across different personality types. For knowledge graph construction, we leverage the role-playing capabilities of large language models (LLMs) to perform reasoning tasks. To enhance the quality of the generated knowledge, we incorporate a debate mechanism consisting of a proponent, an opponent, and a judge, which iteratively refines the outputs through feedback loops. We evaluate the dataset from multiple perspectives and conduct fine-tuning and ablation experiments using multiple LLM backbones to assess PCoKG's robustness and the effectiveness of its construction pipeline. Our LoRA-based fine-tuning results indicate a positive correlation between model performance and the parameter scale of the base models. Finally, we apply PCoKG to persona-based dialogue generation, where it demonstrates improved consistency between generated responses and reference outputs. This work bridges the gap between commonsense reasoning and individual cognitive differences, enabling the development of more personalized and context-aware AI systems.
\end{abstract}

\begin{links}
    \link{Code}{https://github.com/silverbeats/pcs_v2}
\end{links}

\section{Introduction}
\label{sec:introduction}

Commonsense reasoning remains a key challenge in machine intelligence \cite{storks2019commonsense}. To advance this capability, NLP research has developed datasets such as ATOMIC \cite{sap2019atomic,hwang2021comet}, which focus on if-then reasoning about events, including their causes (\textit{xIntent}) and effects (\textit{xEffect}). COMET \cite{bosselut2019comet}, built on ATOMIC, has been applied to emotion recognition \cite{zhao2022cauain}, empathetic dialogue generation \cite{wang2022empathetic,sabour2022cem,tu2022misc}, where contextual understanding is essential.

\begin{figure}[!t]
\centering
\includegraphics[width=\linewidth]{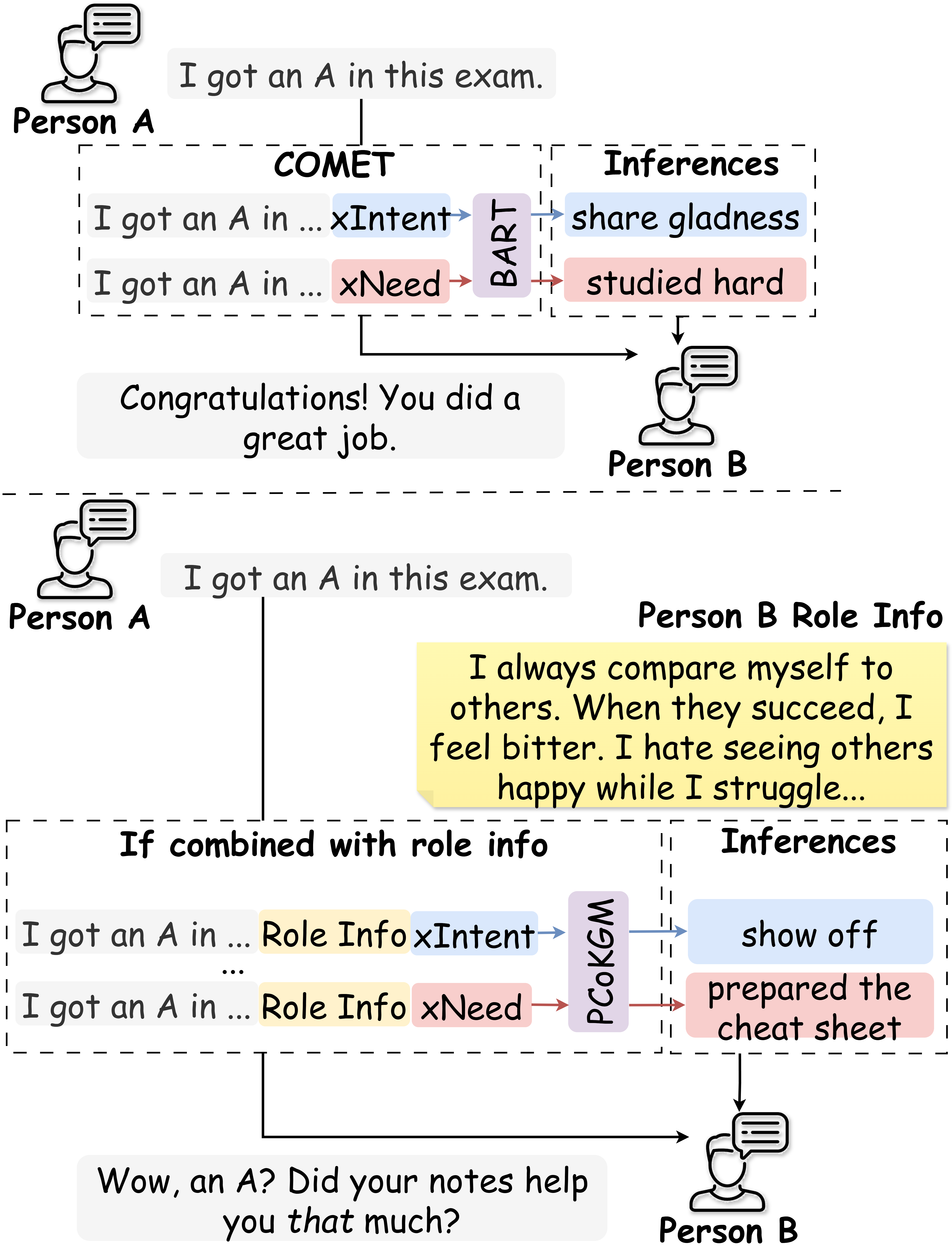}
\caption{The upper panel shows how COMET is commonly used, while the lower panel illustrates a more realistic application incorporating personal traits.}
\label{fig: advertise}
\end{figure}

Commonsense knowledge in ATOMIC is structured as $(e,r,t)$ triplets, where $e$ represents an event, $r$ an inference dimension, and $t$ the outcome. While useful for general reasoning, this framework overlooks individual differences—particularly personality traits—that shape how people interpret and respond to events. As a result, it fails to model real-world cognitive diversity and limits its applicability in personalized AI. As illustrated in the upper part of \myfig{fig: advertise}, existing approaches like COMET generate plausible but generic inferences without considering personal traits. In contrast, the lower part illustrates the significant influence of individual perspectives, resulting in more diverse and personalized responses. Current frameworks, therefore, fall short in capturing the richness of human cognition.

To address this limitation and model individual differences in commonsense reasoning, we introduce the Personality-aware Commonsense Knowledge Graph (PCoKG), which enhances traditional knowledge graphs by integrating person-specific personality traits. We expand the conventional $(e,r,t)$ triplet structure into a quadruple format: $(e,p,r,t)$, where $p$ represents personality information based on the Myers-Briggs Type Indicator (MBTI) \cite{myers1987introduction}. This design enables more realistic modeling of how individuals with different personality types perceive and respond to the same event. For example, an introverted person may view social interactions very differently compared to an extroverted individual. By incorporating personality, our approach facilitates more personalized and context-aware inferences, enabling AI systems better to reflect the cognitive diversity inherent in human reasoning.

Constructing a high-quality commonsense knowledge graph enriched with personality information is challenging. Traditional crowdsourcing methods are limited by the difficulty and cost of recruiting individuals with specific personality traits at scale. To address this, we propose a novel LLM-driven approach that simulates diverse personality types using the MBTI framework. Our method introduces two key components: an evaluator-based mechanism that filters events and inference dimensions to ensure relevance and diversity, and a role-playing prompting strategy enhanced with a multi-agent debate mechanism to improve reasoning consistency and quality. Based on this approach, we construct PCoKG — the first large-scale, personality-aware commonsense knowledge graph, which contains $521,316$ high-quality $(e,p,r,t)$ quadruples across $9$ inference dimensions and all $16$ MBTI types.

Compared to existing resources, PCoKG offers three major advantages. First, it is the first large-scale dataset and modeling framework that explicitly incorporates personality into commonsense reasoning, enabling personalized and contextually appropriate inferences. Second, our LLM-based pipeline eliminates the need for labour-intensive human annotation, supporting scalable data expansion. Third, although the dataset is built upon the MBTI framework, the pipeline is designed to be generalizable, allowing adaptation to other personality theories or the incorporation of additional role attributes. To validate the effectiveness of our approach, we conducted comprehensive experiments using multiple LLM backbones, demonstrating the robustness and generalisation of the generated knowledge across different architectures. We further performed extensive ablation studies to assess the contributions of each component in our pipeline and analyzed performance across model sizes to understand the impact of parameter scale. Finally, we applied PCoKG to a persona-based dialogue generation task, showcasing its practical utility in real-world applications. These results validate both the effectiveness of our methodology and the value of incorporating personality into commonsense knowledge modeling.

\section{Related Work}
\label{sec: related work}

\subsection{Commonsense Knowledge Bases}
ConceptNet \cite{speer2017conceptnet} is a widely used commonsense knowledge base (Commonsense Knowledge Base, CKB), though its Chinese version contains relatively limited knowledge \cite{kuo2009community}. To address such limitations, \citet{zhang2021transomcs} automatically constructed a large-scale commonsense knowledge base called TransOMCS by transforming syntactic parses of web sentences into structured triples. However, most existing CKBs primarily focus on taxonomic relationships, such as \textit{isA} and Synonym \cite{davis2015commonsense}, which inevitably restricts their applicability. In contrast, another line of commonsense knowledge bases—represented by ATOMIC—requires human annotators to infer the causes and consequences of given events based on personal commonsense knowledge, thereby enriching the knowledge graph with tail entities \cite{sap2019atomic,hwang2021comet}. 

Inspired by ATOMIC's organisational structure, \citet{li-etal-2022-c3kg} introduced C$^3$KG, the first Chinese conversational commonsense knowledge graph built upon four types of conversational flows linking headers (events) and tails (inference results). Following the ATOMIC framework, \citet{wang-etal-2024-ecok} built an emotional commonsense knowledge graph. \citet{li2024cpcur} constructed a Chinese dataset for personalized-aware commonsense reasoning, which is the most closely related to our work. However, their dataset was built manually, making it difficult to scale up in terms of size and coverage.

\subsection{Myers-Briggs Type Indicator}
The Myers-Briggs Type Indicator~(MBTI) \cite{myers1987introduction} is a widely recognized personality model that categorizes individuals along four dichotomous dimensions: Introversion/Extraversion, Sensing/Intuition, Thinking/Feeling, and Judging/Perceiving. Despite psychometric critiques \cite{barbuto1997critique}, it is widely used in both professional and personal contexts.

Due to its intuitive appeal, the MBTI has gained traction in computational research. Recent studies have leveraged MBTI for personality-aware natural language processing and human-computer interaction. Examples include MBTI-labeled Reddit datasets \cite{gjurkovic-snajder-2018-reddit}, personalized emotional support systems \cite{tu2023characterchat}, embedding stable MBTI traits into LLMs \cite{cui2023machine}, simulating MBTI types to evaluate LLM decision-making, affective computing datasets integrating MBTI and emotion \cite{zhou2024achinese}, conversational MBTI and gender inference models \cite{Shahnazari2025who}, investigations into personality-adaptive dialogue agents \cite{cheng2025exploring}, and analyses of MBTI-induced bias in hate speech detection \cite{yuan2025hateful}. Collectively, these studies highlight the potential of MBTI-aware modeling to enhance personalisation, fairness, and behavioural fidelity in AI systems.

While existing commonsense knowledge bases primarily emphasise general causal and taxonomic relationships, and some recent efforts have included personality traits in NLP tasks, our work distinguishes itself by systematically incorporating personality-aware reasoning into a large-scale commonsense knowledge graph. Unlike manually curated datasets, which often have limitations in scale and coverage, we propose a fully scalable framework that leverages LLMs to simulate diverse MBTI personality types in commonsense reasoning tasks.

\section{Construction of the Personality-aware Commonsense Knowledge Graph}
\label{sec: dataset}
\begin{figure*}[tb]
\centering
\includegraphics[width=0.9\linewidth]{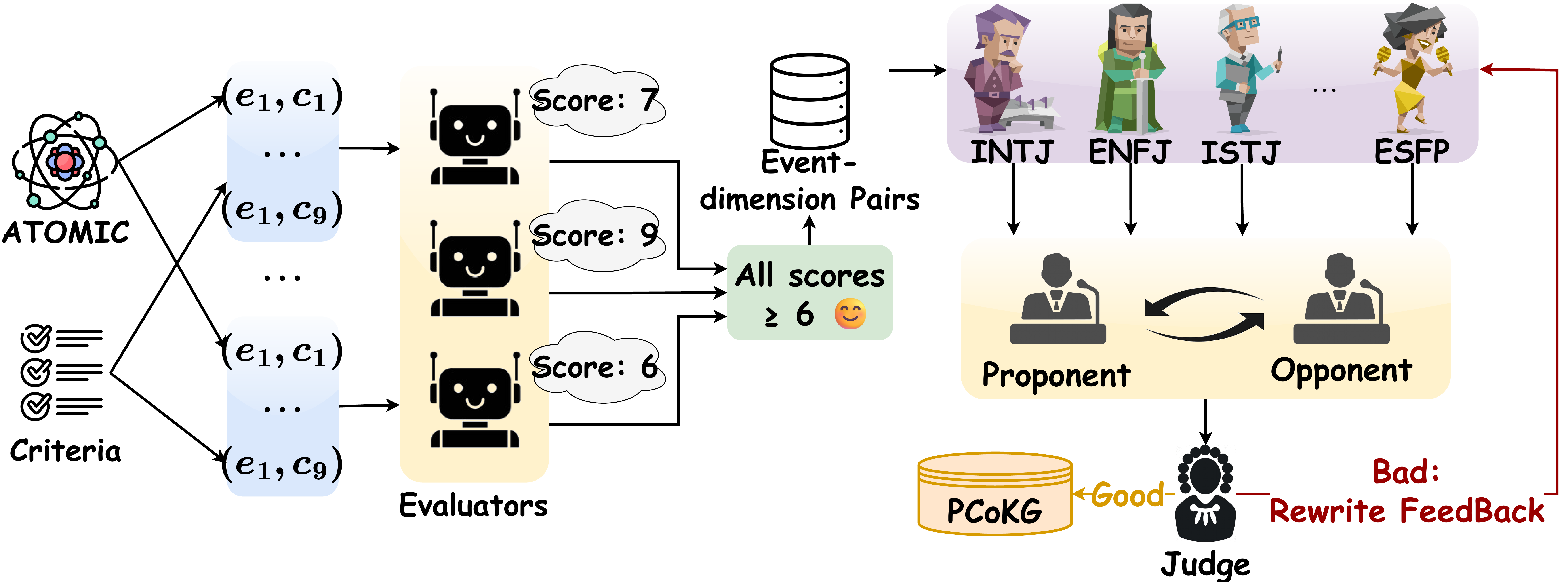}
\caption{PCoKG construction process.}
\label{fig: model}
\end{figure*}

The data construction pipeline consists of two stages, and its key steps are detailed in \myfig{fig: model}.

First, we enumerate event–reasoning pairs $(e, r)$ from the ATOMIC knowledge base and filter them using three LLM evaluators, which score each pair on a 10-point scale for its potential to elicit personality-diverse responses. Only pairs scoring above 6 from all evaluators are retained.

Second, we sample MBTI personality types $p$ according to their global population distribution\footnote{https://www.16personalities.com/country-profiles/global/world\#global} and prompt an LLM to role-play each selected $(e, r)$ pair, generating personalized reasoning outputs $t$. Each instance in PCoKG is thus a quadruple $(e, r, p, t)$ (see examples in \mytable{table: dataset examples}).

The following sections detail these two stages: (1) acquisition of events and reasoning dimensions, and (2) personality-conditioned reasoning generation.

\begin{algorithm}[tb]
\caption{Event and Reasoning Dimension Acquisition}
\label{alg:event_reasoning_extraction_simplified}
\small
\begin{algorithmic}[1]
    \STATE \textbf{Initialize:} 
        $ER \gets \emptyset$ \COMMENT{Set of event-dimension pairs} \\
        $E$ \COMMENT{List of events}, $EVLS$ \COMMENT{List of evaluators}, $CRIT$ \COMMENT{List of criteria}, $CRMap$ \COMMENT{Mapping from criterion to reasoning dimension}

    \FOR{each $e \in E$}
        \FOR{each $c \in CRIT$}
            \STATE $r \gets CRMap(c)$
            \STATE $S \gets \emptyset$
            
            \FOR{each $evl \in EVLS$}
                \STATE $s \gets evl(e, c)$
                \STATE Append $s$ to $S$
            \ENDFOR
            
            \IF{$\forall s \in S,\ s \geq 6$}
                \STATE Append $(e, r)$ to $ER$
            \ENDIF
        \ENDFOR
    \ENDFOR
    \RETURN $ER$
\end{algorithmic}
\end{algorithm}


\begin{table*}[tb]
\centering
\small
\begin{tabularx}{\linewidth}{LllL}
\toprule
Event & Dimension & MBTI & Inference \\ \midrule
PersonX makes any money & xWant & ISFJ & Plan to save some and use the rest for family and friends \\ 
 &  & INTP & Dive deeper into the theoretical aspects to uncover underlying principles \\ \midrule
PersonX locks PersonY's keys in PersonY's car & xIntent & ENFP & Just for fun and to see PersonY's reaction \\ 
& & INTJ & To assess PersonY's response time \\ \bottomrule
\end{tabularx}
\caption{Samples of PCoKG data.}
\label{table: dataset examples}
\end{table*}

\subsection{Event and Reasoning Dimension Acquisition}

\begin{algorithm}[tb]
\caption{Personality-aware Debate Process}
\label{alg: debate_process}
\begin{algorithmic}[1]
    \STATE \textbf{Input:} Event $e$, reasoning dimension $r$, personality type $p$, inference model $M$ 
    \STATE \textbf{Output:} Inference result $t$

    \STATE Initialize empty histories: $rp\_hist$, $pro\_hist$, $opp\_hist$, $jdg\_hist$
            
    \FOR{$i = 1$ to max\_generate\_times}
        \STATE $t \gets M(e, r, p)$

        \FOR{$round = 1$ to debate\_rounds}
            \STATE $pro\_resp \gets \textsc{Pro}(e, r, p, t, pro\_hist)$
            \STATE Append $pro\_resp$ to $pro\_hist$ and $opp\_hist$

            \STATE $opp\_resp \gets \textsc{Opp}(e, r, p, t, opp\_hist)$
            \STATE Append $opp\_resp$ to $pro\_hist$ and $opp\_hist$

            \STATE Append both responses to $jdg\_hist$
        \ENDFOR

        \STATE $jdg\_resp \gets \textsc{Jdg}(e, r, p, jdg\_hist)$

        \IF{$jdg\_resp$ is acceptable}
            \RETURN $t$
        \ELSE
            \STATE Append $jdg\_resp$ to $rp\_hist$ for feedback
        \ENDIF
    \ENDFOR
    \RETURN None
\end{algorithmic}
\end{algorithm}

Initially, events are extracted from ATOMIC$_{20}^{20}$ \cite{hwang2021comet}. To ensure linguistic quality, we employ \textit{language\_tool\_python} to filter out events containing grammatical errors, resulting in a final set of $19,184$ well-formed events. Subsequently, we select three large language models—Deepseek-R1, Qwen-Turbo, and Doubao-1.6-Seed—as evaluation models. We define nine evaluation criteria centred on the reasoning dimension, aiming to assess whether each event applies to reasoning processes associated with different personality types. Each criterion is rated on a $10$-point scale. The overall extraction algorithm is presented in Algorithm~\ref{alg:event_reasoning_extraction_simplified}. The specific evaluation criteria are as follows. Finally, we obtain $95,783$ pairs of $(e, r)$, involving $15,227$ events, denoted as $\mathcal{E}$.

\begin{itemize}
    \item \textbf{xIntent (Motivation)} Does the event lead to clearly different internal drives depending on the character's MBTI type?
    
    \item \textbf{xWant (Plan)} Do different MBTI types form different plans or intentions in response to the event?
    
    \item \textbf{xEffect (Impact)} Does the event have different psychological or behavioural impacts on different MBTI types?
    
    \item \textbf{xReact (Emotional Response)} Do different MBTI types show different emotional reactions to the event?
    
    \item \textbf{xNeed (Preparation)} Would different MBTI types prepare differently for this event?
    
    \item \textbf{xAttr (Self-Narration)} How would different MBTI types describe this event in first person? Is there variation in tone or perspective?
    
    \item \textbf{oReact (Others' Emotion)} Do different MBTI types make different assumptions about how others feel about the event?
    
    \item \textbf{oWant (Others' Intention)} Do different MBTI types expect others to react differently to the event?
    
    \item \textbf{oEffect (Impact on Others)} Do different MBTI types assume different levels of impact on others due to the event?
\end{itemize}

\subsection{Personality-aware Reasoning via Debate}

We guide large language models to adopt roles corresponding to different MBTI personality types and perform reasoning based on the entity-reasoning set ($\mathcal{E}$) obtained in the previous section. To enhance model comprehension and the quality of inferences, we convert each reasoning dimension into clear, human-readable natural language descriptions. 

To further improve the reliability of the generated inferences, we introduce a debate framework, with the core algorithm detailed in Algorithm~\ref{alg: debate_process}. In this framework, we define three roles: \textbf{Proponent}, \textbf{Opponent}, and \textbf{Judge}. The Proponent argues that the model's reasoning aligns with the target MBTI type and provides supporting evidence. In contrast, the Opponent challenges the consistency between the reasoning and the expected type. 

Multiple rounds of debate are held between the Proponent and Opponent, after which the Judge evaluates their arguments and delivers a final judgment. If the initial reasoning does not meet the desired standard, the Judge offers feedback and suggestions for improvement, prompting the model to refine its output iteratively.

\section{Dataset Analysis}

\begin{table}[tb]
    \centering
    \begin{tabularx}{\linewidth}{lC}
        \toprule
        Metric & Value \\ \midrule
        Data Size & 521,316 \\
        Number of Events & 15,077 \\ 
        Average Event Length & 4.79 \\
        Average Reasoning Outcome & 8.75 \\ \bottomrule
    \end{tabularx}
    \caption{Statistical summary of PCoKG.}
    \label{table: stat pcokg}
\end{table}


The basic statistical information of PCoKG is shown in \mytable{table: stat pcokg}. To assess whether the LLM-generated reasoning responses are meaningfully aligned with the assigned MBTI types, we evaluate the dataset through three lenses: Readability-Personality Association, Personality-Reasoning Association, and Human Evaluation.

\subsection{Readability-Personality Association Analysis}

Personality traits are known to influence language use and communication styles. This study uses the Flesch Reading Ease score—a standard measure of text readability (ranging from $0$ to $100$, with higher scores indicating simpler language)—to evaluate linguistic complexity across MBTI types.

\begin{figure}[tb]
\centering
\includegraphics[width=\linewidth]{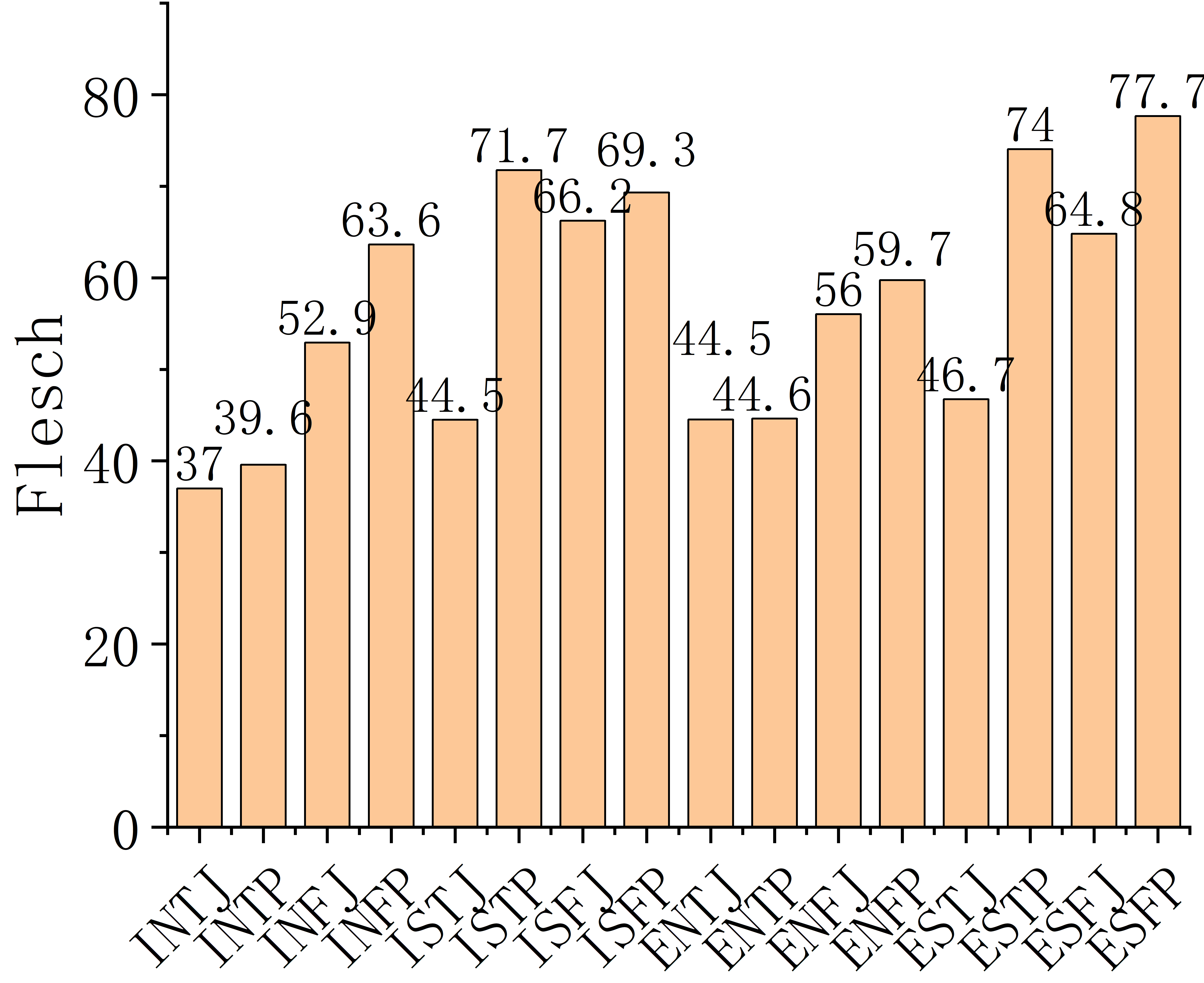}
\caption{Flesch reading ease scores across 16 MBTI personality types.}
\label{fig: flesch}
\end{figure}

\myfig{fig: flesch} shows notable differences in readability across MBTI types. For example, ESFP ($77.7$) and ESTP ($74.0$) have high readability scores, suggesting a preference for direct, concrete expression. In contrast, INTJ ($37.0$) and INTP ($39.6$) score much lower, reflecting more complex and abstract language. These findings align with MBTI theory: Thinking (T) and Intuitive (N) types favour logical and abstract reasoning, while Feeling (F) and Sensing (S) types prefer emotional and accessible language.

Overall, the results suggest that LLM-generated reasoning content reflects linguistic traits consistent with the target MBTI type, supporting its effectiveness in personality perception tasks.

\subsection{Personality-Reasoning Association Analysis}

\begin{table}[tb]
\centering
\small
\begin{tabularx}{\linewidth}{lCC}
\toprule
Dimension & AMI & AMI Shuffle \\ \midrule
xReact & 0.255950 & -0.000009 \\
xWant & 0.237505 & 0.000021 \\
xIntent & 0.240094 & -0.000033 \\
xNeed & 0.177847 & -0.000029 \\
xEffect & 0.216289 & -0.000026 \\
xAttr & 0.511176 & -0.000027 \\
oWant & 0.121604 & 0.000062 \\
oEffect & 0.161030 & -0.000008 \\
oReact & 0.114518 & -0.000044 \\ \bottomrule
\end{tabularx}
\caption{Adjusted MI (AMI) between MBTI personality types and clustering labels of reasoning results ($p\text{-value} <$ 0.01).}
\label{table: ami}
\end{table}

To evaluate whether LLM-generated reasoning reflects meaningful MBTI-aligned signals, we perform mutual information analysis. Reasoning styles are derived via K-means clustering on sentence embeddings, with cluster numbers selected based on Silhouette Scores. We compute Adjusted Mutual Information (AMI) between cluster IDs and MBTI types, which accounts for chance agreement and offers a reliable comparison. As a baseline, we shuffle MBTI labels $100$ times and recalculate AMI, yielding near-zero null values \cite{ojala2002multiresolution}. Mann-Whitney U tests show all original AMI values are significantly higher ($p < 0.01$), confirming strong associations. Results (\mytable{table: ami}) indicate that dimensions related to self-perception (e.g., \textit{xAttr}, \textit{xIntent}) align more clearly with MBTI types, suggesting that LLM-generated reasoning varies meaningfully across personality types.

\subsection{Human Evaluation}

\begin{table}[tb]
\centering
\small
\begin{tabularx}{\linewidth}{lC}
\toprule
Metric & Average Score \\ \midrule
Personality Consistency & 1.63 \\
Reasoning Coherence & 1.78 \\
Naturalness & 1.71 \\ \bottomrule
\end{tabularx}
\caption{Human evaluation results.}
\label{table: human evaluation}
\end{table}

To further validate the presence of personality-aware reasoning signals in our dataset, we conducted a human evaluation in which three psychology graduate students familiar with MBTI assessed $1,440$ randomly selected instances covering all $16$ MBTI types and $9$ reasoning dimensions. Each reasoning result received ratings on three criteria: personality consistency, reasoning coherence, and naturalness, using a $3$-point scale where $0$ indicates \textit{No}, $1$ indicates \textit{Somewhat}, and $2$ indicates \textit{Yes}. The average inter-annotator agreement, measured by Fleiss' Kappa, was $0.57$, indicating moderate to substantial agreement. The results summarized in \mytable{table: human evaluation} indicated that the reasoning responses were generally coherent, with an average score of $1.78$, and natural, with an average score of $1.71$. However, the average score for personality consistency was slightly lower at $1.63$. This suggested that while most responses aligned with the intended personality type, there was still room for improvement.

\section{Experiments}

\subsection{Setup}

\subsubsection{Dataset}
To evaluate the effectiveness of the PCoKG dataset in supporting personalized commonsense reasoning, we partition the data into training, validation, and test sets using a $9$:$0.5$:$0.5$ split based on events. This resulted in $13,576$ training events ($468,479$ quadruples), $750$ validation events ($26,321$ quadruples), and $751$ test events ($26,516$ quadruples).
\subsubsection{Metrics}
We adopt word-overlap metrics—specifically, BLEU-4~(\textbf{B-4}), Rouge-1~(\textbf{R-1}), Rouge-2~(\textbf{R-2}), and Rouge-L~(\textbf{R-L})—to assess the degree of lexical alignment between the generated outputs and the reference answers. These metrics provide a quantitative measure of how well the model reproduces content that resembles the expected responses in terms of n-gram overlap.

\subsubsection{Models}
We select three base models, Qwen3-0.6B, LLaMA3-1B, and MiniCPM4-0.5B, and conduct full-parameter fine-tuning on the PCoKG dataset to equip them with personalized commonsense reasoning capabilities. We refer to the resulting fine-tuned models collectively as \textbf{PCoKGM}. For comparison, we also train the \textbf{COMET} model, into which the reasoning dimensions and personality types are incorporated as special tokens in the tokenizer. The input to the model is constructed by concatenating three components: the event, the reasoning dimension, and the personality type, to predict the corresponding reasoning outcome. Additionally, we design one-shot prompting templates to evaluate the performance of Deepseek-R1~(\textbf{R1}), Doubao-seed-1.6-thinking~(\textbf{1.6-Thinking}), and GPT-o4-mini~(\textbf{o4-mini}) on the PCoKG task.

\subsubsection{Implementation Details}
We fine-tune the models using \texttt{LLaMA-Factory} on four 3090 GPUs. Each GPU is assigned a batch size of $8$, with gradient accumulation over $4$ steps. The warmup ratio is set to $0.1$, and the cosine learning rate scheduler is employed. The models are trained for one epoch on the training set, with validation performance evaluated every $300$ training steps. Early stopping is applied if the performance on the validation set does not improve for three consecutive evaluations.

\begin{table}[t]
\centering
\small
\begin{tabularx}{\linewidth}{l*{4}L}
\toprule
Model & B-4 & R-1 & R-2 & R-L \\ \midrule
R1 & 2.67 & 14.45 & 1.89 & 13.44 \\
1.6-Thinking & 3.39 & 15.43 & 2.45 & 13.96 \\
o4-mini & 5.38 & 15.34 & 2.09 & 14.28 \\ \midrule
\textit{COMET} &  &  &  &  \\
- LLaMA3 & 12.58 & 30.51 & 12.77 & 28.91 \\
- Qwen3 & 10.09 & 26.31 & 9.26 & 25.00 \\
- MiniCPM4 & 10.81 & 28.29 & 10.43 & 26.78 \\ \midrule
\textit{PCoKGM} &  & && \\
- LLaMA3 & 13.73 & 32.09 & 14.31 & 30.53 \\
- Qwen3 & 14.08 & 32.68 & 14.78 & 31.07 \\
- MiniCPM4 & \textbf{14.50} & \textbf{32.99} & \textbf{15.27} & \textbf{31.38} \\ \bottomrule
\end{tabularx}
\caption{Comparison of model performance on PCoKG. Best results are indicated in \textbf{bold}.}
\label{table: full sft test result}
\end{table}

\subsection{Comparative Performance Analysis}

As shown in \mytable{table: full sft test result}, PCoKGM significantly outperforms all baseline models across all metrics, achieving the highest scores. This demonstrates that incorporating reasoning dimensions and personality types as natural language prompts into the input enhances the model's ability to generate contextually relevant and personalized commonsense knowledge. In contrast, the COMET model encodes these signals as special tokens rather than interpretable natural language. While this approach enables structured control over reasoning patterns, it may limit interpretability and generalization in downstream applications.

Among the evaluated large language models, the three models exhibit comparable performance, yet all fall short of PCoKGM. This highlights the advantages of domain-specific fine-tuning and the explicit integration of personalization signals into the model's input structure. Moreover, the generated outputs from these LLMs indicate that, although they are capable of role-playing to some extent, their initial generations do not fully align with the desired outcomes. This further validates the necessity of employing a debate framework during PCoKG construction, where generations are refined through feedback from a judge model.

\subsection{Ablation Experiment Analysis}

\begin{table}[htbp]
\centering
\small
\begin{tabularx}{\linewidth}{l*{4}L}
\toprule
Base Model & B-4 & R-1 & R-2 & R-L \\ \midrule
LLaMA3 & \textbf{13.73} & \textbf{32.09} & \textbf{14.31} & \textbf{30.53} \\
- w/o mbti & 10.16&25.59&9.36&24.51 \\
- w/o select &11.25&27.92&10.59&26.49 \\
- w/o debate &12.09 &29.45&12.04&28.08 \\
- w/o select \& debate &10.66&26.00&9.62&24.72 \\ \midrule
Qwen3 & \textbf{14.08} & \textbf{32.68} & \textbf{14.78} & \textbf{31.07} \\
- w/o mbti & 10.33 & 25.72 & 9.55 & 24.68 \\
- w/o select & 10.56 & 28.72 & 9.38 & 24.53 \\
- w/o debate & 11.00 & 26.84 & 9.96 & 25.62 \\
- w/o select \& debate & 10.66 & 26.60 & 9.78 & 25.28 \\ \midrule
MiniCPM4 & \textbf{14.50} & \textbf{32.99} & \textbf{15.27} & \textbf{31.38} \\
- w/o mbti & 10.45 & 25.94 & 9.66 & 24.88 \\
- w/o select & 10.89 & 27.33 & 10.21 & 25.91 \\
- w/o debate & 11.46 & 27.46 & 10.68 & 26.25 \\
- w/o select \& debate & 10.93 & 26.51 & 10.14 & 25.19 \\ \bottomrule
\end{tabularx}
\caption{The ablation experimental results of PCoKGM using different base models. Best results are in \textbf{bold}.}
\label{table: results of ablation experiments}
\end{table}

To further validate the effectiveness of each component in our dataset construction, we design four ablation settings: \textit{w/o mbti}, \textit{w/o select}, \textit{w/o debate}, and \textit{w/o select \& debate}. The results, summarized in \mytable{table: results of ablation experiments}, highlight the distinct contribution of each module.

\begin{figure*}[!t]
\centering
\includegraphics[width=\linewidth]{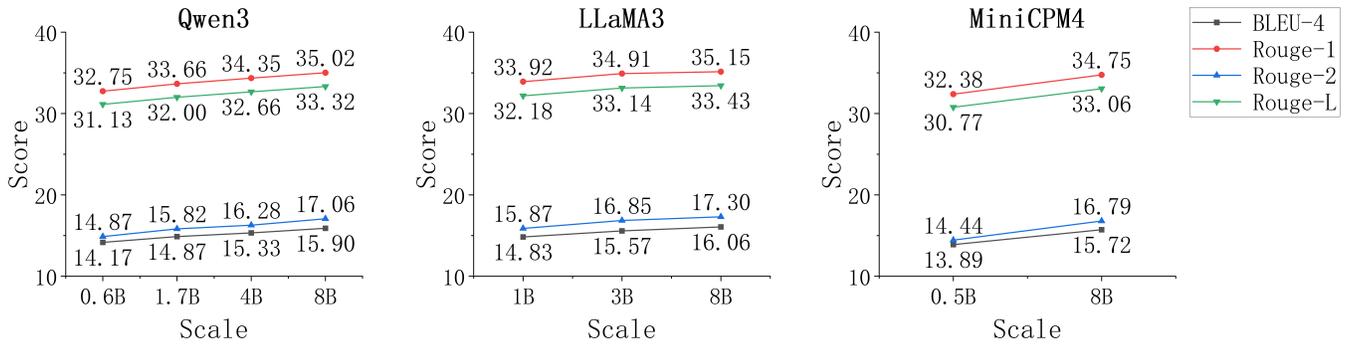}
\caption{Performance of the foundation model with varying sizes on the PCoKG under LoRA fine-tuning.}
\label{fig: impact of model scale lora sft}
\end{figure*}

Removing MBTI personality types (\textit{w/o mbti}) leads to the largest performance drop across all models and metrics, underscoring their critical role in structured reasoning. Removing the selection mechanism (\textit{w/o select}) also reduces performance, indicating that while learning from raw data is possible, curated high-quality event–dimension pairs enhance reasoning accuracy. The \textit{w/o debate} setting results in a moderate decline, suggesting that the debate framework enhances reasoning depth and ensures consistency in personality-aligned responses. The worst performance occurs when both selection and debate are removed (\textit{w/o select \& debate}), confirming their complementary roles.

In summary, each component plays a distinct role: MBTI provides the reasoning anchor, selection ensures data quality, and debate enhances the consistency of personality-aware reasoning. These findings highlight the value of a structured, multi-stage dataset construction process for personality-driven reasoning tasks.

\subsection{Impact of Model Scale}

We evaluate three model families—Qwen3, LLaMA3, and MiniCPM4—across multiple scales to study how model size affects performance on our personality-aware commonsense reasoning dataset. As shown in \myfig{fig: impact of model scale lora sft}, performance improves with scale across all models, suggesting that larger models better capture the nuanced relationships between personality types, events, and reasoning outcomes. This trend is consistent across BLEU-4 and ROUGE metrics, indicating enhanced fluency and coherence in generated inferences as model capacity increases. Notably, the performance gap between model families diminishes at larger scales, implying that sufficient model size may compensate for architectural or training differences. Additionally, improvements in ROUGE-1, ROUGE-2, and ROUGE-L largely align with the BLEU-4 trends, reinforcing the notion that larger models are better equipped to handle complex, personality-conditioned reasoning tasks. These findings underscore the significance of model scale in discerning fine-grained behavioral patterns from large-scale structured data. Overall, the results suggest that scaling up model size is a viable strategy for improving reasoning accuracy and coherence, especially when dealing with multi-dimensional, personality-sensitive inference tasks.

\subsection{Application}

To validate the effectiveness of our proposed PCoKG in practical applications, we conduct experiments on a dialogue generation task. Specifically, we utilize the SPC dataset's test set (967 samples) \cite{jandaghi-etal-2024-faithful} to evaluate how well our approach enhances personalized reasoning compared to existing methods.

Initially, we trained a BERT classifier on the \textit{MBTI Personality Types 500 Dataset}\footnote{https://www.kaggle.com/datasets/zeyadkhalid/mbti-personality-types-500-dataset} from Kaggle. This classifier was then employed to predict the MBTI personality types of the conversational participants in the SPC dataset based on their character descriptions. Using these predictions, PCoKGM—fully fine-tuned on MiniCPM4-0.5B—generates responses that integrate context and personality. For comparison, we also evaluate COMET, which relies solely on contextual content for inference.

The results, summarized in \mytable{table: application}, reveal several key insights. First, integrating commonsense reasoning into dialogue generation yields better outcomes. Both COMET and PCoKGM outperform models without commonsense reasoning across various metrics, indicating that leveraging commonsense knowledge significantly enhances the alignment between generated outputs and expected responses. Second, incorporating personality traits into commonsense reasoning further improves the quality of generated dialogues. Across all tested models, PCoKGM consistently outperforms COMET. These findings suggest that modeling personality-specific nuances leads to more accurate and engaging dialogues.

\begin{table}[tb]
\centering
\small
\begin{tabularx}{\linewidth}{ll*{4}L}
\toprule
Model & B-4 & R-1 & R-2 & R-L \\ \midrule
R1 & 10.40 & 26.04 & 10.88 & 24.88 \\
- w/ COMET & 11.76 & 28.58 & 13.24 & 27.40 \\
- w/ PCoKGM & \textbf{12.03} & \textbf{29.69} & \textbf{13.89} & \textbf{28.47} \\ \midrule
1.6-Thinking & 16.51 & 37.41 & 20.17 & 35.80 \\
- w/ COMET & 18.28 & 40.31 & 22.19 & 38.92 \\
- w/ PCoKGM & \textbf{18.92} & \textbf{40.89} & \textbf{22.85} & \textbf{39.50} \\ \midrule
o4-mini & 6.05 & 18.55 & 5.69 & 17.63 \\
- w/ COMET & 6.65 & 20.24 & 6.84 & 19.16 \\
- w/ PCoKGM & \textbf{6.85} & \textbf{20.27} & \textbf{7.09} & \textbf{19.31} \\ \bottomrule
\end{tabularx}
\caption{Comparison of dialogue generation performance. Best results are in \textbf{Bold}.}
\label{table: application}
\end{table}

\section{Conclusion}
This study introduces Personality-aware Commonsense Knowledge Graph (PCoKG), a knowledge graph enhanced with personality traits to support personalized, context-aware reasoning. It integrates event quality control, reasoning dimensions, and a role-based debate mechanism to build a large-scale knowledge graph. Experiments confirm the effectiveness of each component and explore the impact of model scale. We also demonstrate PCoKG's value in dialogue generation. Limitations include a focus on personality traits alone, without considering factors like gender or occupation. Future work will incorporate these aspects to develop a more comprehensive framework for personalized reasoning.

\section*{Acknowledgments}
This work was supported by the National Natural Science Foundation of China (No. 62376178), Jiangsu Key Laboratory of Language Computing(JSLCKeyLab 202500003), and Project Funded by the Priority Academic Program Development of Jiangsu Higher Education Institutions.

\bibliography{aaai2026}

@misc{storks2019commonsense,
    author = {{Storks}, Shane and {Gao}, Qiaozi and {Chai}, Joyce Y.},
    title = "{Recent Advances in Natural Language Inference: A Survey of Benchmarks, Resources, and Approaches}",
    year = 2019,
    archivePrefix = {arXiv},
    eprint = {1904.01172},
}

@InProceedings{sap2019atomic,
  author    = {Sap, Maarten and Le Bras, Ronan and Allaway, Emily and Bhagavatula, Chandra and Lourie, Nicholas and Rashkin, Hannah and Roof, Brendan and Smith, Noah A and Choi, Yejin},
  booktitle = {Proc. AAAI},
  title     = {Atomic: An atlas of machine commonsense for if-then reasoning},
  year      = {2019},
  pages     = {3027--3035},
  publisher={AAAI Press}
}

@InProceedings{hwang2021comet,
    author    = {Hwang, Jena D and Bhagavatula, Chandra and Le Bras, Ronan and Da, Jeff and Sakaguchi, Keisuke and Bosselut, Antoine and Choi, Yejin},
    booktitle = {Proc. AAAI},
    title     = {(Comet-) atomic 2020: on symbolic and neural commonsense knowledge graphs},
    year      = {2021},
    pages     = {6384--6392},
    address={Virtual Event},
    publisher={AAAI Press}
}

@InProceedings{zhao2022cauain,
    author    = {Zhao, Weixiang and Zhao, Yanyan and Lu, Xin},
    booktitle = {Proc. IJCAI},
    title     = {Cauain: Causal aware interaction network for emotion recognition in conversations},
    year      = {2022},
    pages     = {4524--4530},
    address={Vienna, Austria},
    publisher={IJCAI Organization},
}

@inproceedings{wang2022empathetic,
    author  = {Lanrui Wang and
    Jiangnan Li and
    Zheng Lin and
    Fandong Meng and
    Chenxu Yang and
    Weiping Wang and
    Jie Zhou},
    booktitle    = {Proc. Findings of EMNLP},
    title        = {Empathetic Dialogue Generation via Sensitive Emotion Recognition and Sensible Knowledge Selection},
    year         = {2022},
    pages        = {4634--4645},
    address = {Abu Dhabi, United Arab Emirates},
    publisher    = {ACL},
}

@InProceedings{sabour2022cem,
    author    = {Sabour, Sahand and Zheng, Chujie and Huang, Minlie},
    booktitle = {Proc. AAAI},
    title     = {Cem: Commonsense-aware empathetic response generation},
    year      = {2022},
    pages     = {11229--11237},
    address={Virtual Event},
    publisher    = {AAAI Press},
}

@InProceedings{tu2022misc,
    author    = {Tu, Quan and Li, Yanran and Cui, Jianwei and Wang, Bin and Wen, Ji-Rong and Yan, Rui},
    booktitle = {Proc. ACL},
    title     = {MISC: A Mixed Strategy-Aware Model integrating COMET for Emotional Support Conversation},
    year      = {2022},
    pages     = {308--319},
    address={Dublin, Ireland},
    publisher={ACL}
}

@InProceedings{bosselut2019comet,
  author    = {Bosselut, Antoine and Rashkin, Hannah and Sap, Maarten and Malaviya, Chaitanya and Celikyilmaz, Asli and Choi, Yejin},
  booktitle = {Proc. ACL},
  title     = {{COMET}: Commonsense Transformers for Automatic Knowledge Graph Construction},
  year      = {2019},
  pages     = {4762--4779},
  address   = {Florence, Italy},
  publisher = {ACL}
}

@InProceedings{li2024cpcur,
    author="Yang, Yong
    and Li, Weijie
    and Fan, Xiaochao
    and Deng, Wenjun
    and Liu, Jiapeng
    and Diao, Yufeng
    and Tuerxun, Palidan",
    booktitle="Proc. NLPCC",
    title="Chinese Personalized Commonsense Understanding and Reasoning Based on Curriculum-Learning",
    year="2024",
    pages="213--225",
    address="Hangzhou, China",
    publisher="Springer Nature Singapore",
}

@InProceedings{speer2017conceptnet,
  author    = {Speer, Robyn and Chin, Joshua and Havasi, Catherine},
  booktitle = {Proc. AAAI},
  title     = {Conceptnet 5.5: An open multilingual graph of general knowledge},
  year      = {2017},
  pages     = {4444--4451},
  address   = {San Francisco, California, USA},
  publisher = {AAAI Press},
}

@InProceedings{kuo2009community,
  author    = {Kuo, Yen-ling and Lee, Jong-Chuan and Chiang, Kai-yang and Wang, Rex and Shen, Edward and Chan, Cheng-wei and Hsu, Jane Yung-jen},
  booktitle = {Proc. ACM SIGKDD Workshop on Human Computation},
  title     = {Community-Based Game Design: Experiments on Social Games for Commonsense Data Collection},
  year      = {2009},
  pages     = {15--22},
  address   = {Paris, France},
  publisher = {ACM},
}

@InProceedings{zhang2021transomcs,
  author    = {Zhang, Hongming and Khashabi, Daniel and Song, Yangqiu and Roth, Dan},
  booktitle = {Proc. IJCAI},
  title     = {TransOMCS: from linguistic graphs to commonsense knowledge},
  year      = {2021},
  address   = {Virtual Event},
  pages     = {4004--4010},
  publisher = {IJCAI Organization},
}

@Article{davis2015commonsense,
  author     = {Davis, Ernest and Marcus, Gary},
  journal    = {Commun. ACM},
  title      = {Commonsense Reasoning and Commonsense Knowledge in Artificial Intelligence},
  year       = {2015},
  month      = {aug},
  number     = {9},
  pages      = {92--103},
  volume     = {58},
  address    = {New York, NY, USA},
  numpages   = {12},
  publisher  = {ACM}
}

@InProceedings{li-etal-2022-c3kg,
  author    = {Li, Dawei and Li, Yanran and Zhang, Jiayi and Li, Ke and Wei, Chen and Cui, Jianwei and Wang, Bin},
  booktitle = {Proc. Findings of ACL},
  title     = {{C}$^3${KG}: A {C}hinese Commonsense Conversation Knowledge Graph},
  year      = {2022},
  pages     = {1369--1383},
  address   = {Dublin, Ireland},
  publisher = {ACL},
}

@inproceedings{wang-etal-2024-ecok,
    author = "Wang, Zhunheng  and
      Liu, Xiaoyi  and
      Hu, Mengting  and
      Ying, Rui  and
      Jiang, Ming  and
      Wu, Jianfeng  and
      Xie, Yalan  and
      Gao, Hang  and
      Cheng, Renhong",
    booktitle = "Proc. Findings of ACL",
    title = "{EC}o{K}: Emotional Commonsense Knowledge Graph for Mining Emotional Gold",
    year = "2024",
    pages = "8055--8074",
    address = "Bangkok, Thailand",
    publisher = "ACL",
}

@article{barbuto1997critique,
  title={A critique of the Myers-Briggs Type Indicator and its operationalization of Carl Jung's psychological types},
  author={Barbuto Jr, John E},
  journal={Psychological Reports},
  volume={80},
  number={2},
  pages={611--625},
  year={1997},
  publisher={SAGE Publications Sage CA: Los Angeles, CA}
}

@book{myers1987introduction,
  title={Introduction to type: A description of the theory and applications of the Myers-Briggs Type Indicator},
  author={Myers, Isabel Briggs},
  year={1987},
  publisher={Consulting Psychologists Press}
}

@inproceedings{gjurkovic-snajder-2018-reddit,
    title = "{R}eddit: A Gold Mine for Personality Prediction",
    author = "Gjurkovi{\'c}, Matej  and
      {\v{S}}najder, Jan",
    booktitle = "Proc. Workshop on PEOPLES",
    month = jun,
    year = "2018",
    address = "New Orleans, Louisiana, USA",
    publisher = "ACL",
    pages = "87--97"
}

@misc{cui2023machine,
    title={Machine mindset: An mbti exploration of large language models},
    author={Cui, Jiaxi and Lv, Liuzhenghao and Wen, Jing and Wang, Rongsheng and Tang, Jing and Tian, Yonghong and Yuan, Li},
    year={2023},
    archivePrefix = {arXiv},
    eprint = {2312.12999},
}

@misc{tu2023characterchat,
  title={Characterchat: Learning towards conversational ai with personalized social support},
  author={Tu, Quan and Chen, Chuanqi and Li, Jinpeng and Li, Yanran and Shang, Shuo and Zhao, Dongyan and Wang, Ran and Yan, Rui},
  year={2023},
archivePrefix = {arXiv},
    eprint = {2308.10278},
}

@misc{zhou2024achinese,
       author = {{Zhou}, Jingyi and {Luo}, Senlin and {Chen}, Haofan},
        title = "{A Chinese Multi-label Affective Computing Dataset Based on Social Media Network Users}",
         year = {2024},
archivePrefix = {arXiv},
    eprint = {2411.08347}
}

@misc{Shahnazari2025who,
       author = {{Shahnazari}, Kourosh and {Moein Ayyoubzadeh}, Seyed},
        title = "{Who Are You Behind the Screen? Implicit MBTI and Gender Detection Using Artificial Intelligence}",
         year = {2025},
archivePrefix = {arXiv},
    eprint = {2503.09853}
}

@misc{cheng2025exploring,
       author = {{Cheng}, Sijia and {Chang}, Wen-Yu and {Chen}, Yun-Nung},
        title = "{Exploring Personality-Aware Interactions in Salesperson Dialogue Agents}",
         year = {2025},
archivePrefix = {arXiv},
    eprint = {2504.18058}
}

@misc{yuan2025hateful,
       author = {{Yuan}, Shuzhou and {Nie}, Ercong and {Tawfelis}, Mario and {Schmid}, Helmut and {Sch{\"u}tze}, Hinrich and {F{\"a}rber}, Michael},
        title = "{Hateful Person or Hateful Model? Investigating the Role of Personas in Hate Speech Detection by Large Language Models}",
         year = {2025},
archivePrefix = {arXiv},
    eprint = {2506.08593}
}

@inproceedings{jandaghi-etal-2024-faithful,
    title = "Faithful Persona-based Conversational Dataset Generation with Large Language Models",
    author = "Jandaghi, Pegah  and
      Sheng, Xianghai  and
      Bai, Xinyi  and
      Pujara, Jay  and
      Sidahmed, Hakim",
    booktitle = "Proc. Workshop on NLP4ConvAI",
    month = aug,
    year = "2024",
    address = "Bangkok, Thailand",
    publisher = "ACL",
    pages = "114--139",
}

@article{ojala2002multiresolution,
  title={Multiresolution gray-scale and rotation invariant texture classification with local binary patterns},
  author={Ojala, Timo and Pietikainen, Matti and Maenpaa, Topi},
  journal={TPAMI},
  volume={24},
  number={7},
  pages={971--987},
  year={2002},
  publisher={IEEE}
}

\end{document}